\documentclass{llncs}

\usepackage[latin1]{inputenc}
\usepackage{makeidx}  

\begin{document}

\mainmatter           

\title{Parameter-less Optimization with the Extended Compact 
	Genetic Algorithm and Iterated Local Search}
\titlerunning{Parameter-less Optimization with ECGA and ILS}

\author{Cláudio Lima \and Fernando Lobo}
\authorrunning{C. Lima \and F. Lobo}   

\institute{ADEEC, FCT\\
           Universidade do Algarve\\
	   Campus de Gambelas\\
           8000-062 Faro, Portugal\\
           \{clima,flobo\}@ualg.pt}

\maketitle

\begin{abstract}
This paper presents a parameter-less optimization framework that uses 
the extended compact genetic algorithm (ECGA) and iterated local 
search (ILS), but is not restricted to these algorithms. The presented 
optimization algorithm (ILS+ECGA) comes as an extension of the 
parameter-less genetic algorithm (GA), where the parameters of a 
selecto-recombinative GA are eliminated. The approach that we propose 
is tested on several well known problems. In the absence of domain 
knowledge, it is shown that ILS+ECGA is a robust and easy-to-use 
optimization method.
\end{abstract}

\section{Introduction}

  One of the major topics of discussion within the evolutionary computation 
  community has been the parameter specification of the evolutionary 
  algorithms (EAs).
  After choosing the encoding and the operators to use, the EA user needs
  to specify a number of parameters that have little to do with the problem
  (from the user perspective), but more with the EA mechanics itself. In order
  to release the user from the task of setting and tuning the EA parameters,
  several techniques have been proposed. One of these techniques is the
  parameter-less GA, which controls the parameters of a
  selecto-recombinative GA. This technique can be applied to various types
  of (selecto-recombinative) GAs, and in conjunction with a high-order 
  estimation of distribution algorithm (EDA), such as the extended compact 
  GA (ECGA) or the Bayesian optimization algorithm (BOA), results in a 
  powerful and easy-to-use search algorithm. Multivariate EDAs have shown 
  to outperform the SGA by 
  several orders of magnitude, especially on very difficult problems. 
  However, these advanced search algorithms don't come for free, requiring 
  more computational effort than the SGA when moving from population 
  to population. In many problems this extra effort is well worth it, but 
  for other (less complex) problems, a simpler algorithm can 
  easily outperform a multivariate EDA.

  Typical EAs are based on two variation operators: recombination and mutation.
  Recombination and mutation search the solution space in different
  ways and with different resources. While recombination needs large
  populations to combine effectively the necessary information, mutation
  works best when applied to small populations during a large number
  of generations. Spears \cite{Spears:93} did a comparative study between 
  crossover and mutation operators, and theoretically demonstrates that 
  there were some important characteristics of each operator that were not 
  captured by the other.

  Based on these observations, we propose a new parameter-less optimization 
  framework, that consists of running two different search models 
  simultaneously. 
  The idea is to use the best of both search strategies in order to obtain 
  an algorithm that works reasonably well in a large class of problems.
  The first method can be a parameter-less ECGA, based on 
  selection and wise recombination to improve a population 
  of solutions. As a second method we can use an iterated local search (ILS) 
  algorithm with adaptive perturbation strength. Instead of working with 
  a population of solutions, the ILS iterates a single solution by means of
  selection and mutation. We called optimization framework, instead of 
  optimization algorithm, since what we propose here is not tied up with 
  the ECGA or our ILS implementation. 
  Other algorithms, such as BOA or other ILS implementations, can be 
  considered with similar or better results. However, in this paper we 
  restrict ourselves to the concrete implementation of the ILS+ECGA 
  algorithm and the discussion of the corresponding results. 

  The next section reviews some of the work done in the topic of EA parameter 
  tuning/control, then describes the parameter-less GA technique, the ECGA, 
  and the ILS framework. Then, Section \ref{sec:ils+ecga} describes the 
  basic principles of the parameter-less optimization framework and our 
  ILS+ECGA implementation. In Section \ref{sec:experiments}, computational 
  experiments are done to validate the proposed approach. Section 
  \ref{sec:extensions} highlights some extensions of this work.
  Finally, in Section \ref{sec:conclusions}, a summary and conclusions are 
  presented.

\section{Related Work} \label{sec:relatedwork}

  This section reviews some of the research efforts done in setting and 
  adapting the EAs parameters, describes the parameter-less GA technique 
  and the mechanics of the ECGA and ILS.

  \subsection{Parameter Tuning and Parameter Control in EAs}

    Parameter tuning in EAs involves the empirical and
    theoretical studies done to find optimal settings and understand the
    interactions between the various parameters. An example of that was the
    work of De Jong \cite{DeJong:75}, where various combinations of 
    parameters were tested on a set of five functions. On those experiments,
    De Jong verified that the parameters that gave better overall 
    performance were: population size in the range 50-100, crossover 
    probability of 0.6, mutation probability of 0.001, and
    generation gap of 1.0 (full replacement of the population in each
    generation). 
    Some other empirical studies have been conducted on a larger set of 
    problems yielding somewhat similar results \cite{Grefenstette:86,Schaffer:89}.
    Almost 30 years later, these parameters are still known as the 
    ``standard'' parameters, being sometimes incorrectly applied to many 
    problems.
    Beside these empirical studies, some work was done to analyze the
    effect of one or two parameters in isolation, ignoring the others.
    Among the most relevant studies, are the ones done on selection
    \cite{Goldberg:91}, population sizing \cite{Goldberg:92,Harik:97a}, 
    mutation \cite{Muhlenbein:92,Back:93a}, and control maps 
    \cite{Goldberg:93,Thierens:93}. The work on population sizing
    is of special relevance, showing that setting the population size to
    50-100 for all problems is a mistake. The control maps study
    gave regions of the parameter space (selection and crossover values)
    where the GA was expected to work well, under the assumption of proper 
    linkage identification.

    In parameter control we are interested in adapting the parameters 
    during the EA run. Parameter control techniques can be sub-divided in 
    three types: deterministic, adaptive, and self-adaptive 
    \cite{Eiben:99}. In deterministic control, the parameters are changed
    according to deterministic rules without using any feedback from the
    search. The adaptive control takes place when there is some form of
    feedback that influences the parameter specification. Examples of
    adaptive control are the works of Davis \cite{Davis:89}, 
    Julstrom \cite{Julstrom:95}, and Smith \& Smuda \cite{Smith:95}.
    The parameter-less GA technique is a mix of deterministic and adaptive 
    rules of control, as we will see in the next section.
    Finally, self-adaptive control is based on the idea that evolution can
    be also applied in the search for good parameter values. In this type
    of control, the operator probabilities are encoded together with the
    corresponding solution, and undergo recombination and mutation. This
    way, the better parameters values will tend to survive because they 
    originate better solutions. Self-adaptive evolution strategies (ESs)
    \cite{Back:95} are an example of the application of this type of 
    parameter control.

  \subsection{Parameter-less Genetic Algorithm} \label{sec:plga}

    The parameter-less genetic algorithm \cite{Harik:99} is a technique
    that eliminates the parameters of a selecto-recombinative GA. Based
    on the schema theorem \cite{Holland:75} and various
    facet-wise theoretical studies of GAs \cite{Harik:97a,Goldberg:93},
    Harik \& Lobo automated the specification of the selection pressure 
    ($s$), crossover rate ($p_c$), and population size ($N$) parameters.

    The selection pressure and crossover rate are set to fixed values,
    according to a simplification of the schema theorem in order to ensure
    the growth of promising building blocks. 
    Simplifying the schema theorem, and under the conservative
    hypothesis that a schema is destroyed during the crossover operation,
    the growth ratio of a schema can be expressed by $s ~ (1 - p_c)$.
    Thus, setting $s=4$ and $p_c=0.5$, gives a net growth factor of 2,
    ensuring that the necessary building blocks will grow. If these
    building blocks will be able to mix in a single individual or not is
    now a matter of having the right population size.

    In order to achieve the right population size, multiple 
    populations with different sizes are run in a concurrent way. 
    The GA starts by firing the first population, with size $N_{1}=4$, 
    and whenever a new population is created its size is doubled. 
    The parameter-less GA gives an advantage to smaller populations 
    by giving them more function evaluations. Consequently, the 
    smaller populations have the chance to converge faster than 
    the large ones. The reader is referred to Harik \& Lobo paper 
    \cite{Harik:99} for details of this approach. 

  \subsection{Extended Compact Genetic Algorithm}

    The extended compact genetic algorithm (ECGA) \cite{Harik:99a}
    is based on the idea that the choice of a good probability
    distribution is equivalent to linkage learning. The ECGA uses
    a product of marginal distributions on a partition of the
    decision variables. These kind of probability distributions are 
    a class of probability models known as marginal product models 
    (MPMs). The measure of a good MPM is quantified based on the
    minimum description length (MDL) principle. According to
    Harik, good distributions are those under which
    the representation of the distribution using the current
    encoding, along with the representation of the population
    compressed under that distribution, is minimal. Formally, the
    MPM complexity is given by the sum $C_m + C_p$. The model 
    complexity $C_m$ is given by

    \begin{equation}
    C_m = \log_{2}(N+1) \sum_{i} (2^{S_{i}} - 1),
    \end{equation}

    \noindent
    where N is the population size and $S_{i}$ is the length of the
    $i^{th}$ subset of genes. The compressed population complexity
    $C_p$ is quantified by

    \begin{equation}
    C_p =  N \sum_{i} {E}(M_{i}),
    \end{equation}

    \noindent
    where ${E}(M_{i})$ is the entropy of the marginal distribution of
    subset $i$. Entropy is a measure of the dispersion (or randomness)
    of a distribution, and is defined as
    $E = \sum_{j=1}^{n} - p_{j} \log_{2}(p_{j})$, where $p_j$ is the 
    probability of observing the outcome $j$ in a total of $n$ possible 
    outcomes.

    \begin{figure}[!t]
    \centering
    \begin{tabular}{@{} p{\textwidth} @{}}
    \hline
    \textbf{\emph{Extended Compact Genetic Algorithm} (ECGA)}\\
        \hline
        \vspace{0.05cm}

    \begin{minipage}[t]{\textwidth}
    \begin{enumerate}
        \item[(1)] Create a random population of $N$ individuals.

    \item[(2)] Apply selection.

    \item[(3)] Model the population using a greedy MPM search.

    \item[(4)] Generate a new population according to the MPM found in
           step 3.

    \item[(5)] If stopping criteria is not satisfied, return to step 2.
    \end{enumerate}
    \end{minipage}\\

    \vspace{0.0cm}\\
    \hline
    \end{tabular}
    \caption{Steps of the extended compact genetic algorithm (ECGA).}
    \label{fig:ecga}
    \end{figure}

    As we can see in Figure \ref{fig:ecga}, steps 3 and 4 of the
    ECGA differ from the simple GA operation. Instead of applying
    crossover and mutation, the ECGA searches for a MPM that better
    represents the current population and then generates a new
    population sampling from the MPM found in step 3. This way, new
    individuals are generated without destroying the building blocks.

  \subsection{Iterated Local Search}

    The iterated local search (ILS) \cite{Lourenco:02} is a simple
    and general purpose meta-heuristic that iteratively builds a
    sequence of solutions generated by an embedded heuristic,
    leading to better solutions than repeated random
    trials of that heuristic. This simple idea is not new, but
    Lourenço et al. formulated as a general framework.
    The key idea of ILS is to build a biased randomized walk in the
    space of local optima, defined by some local search algorithm.
    This walk is done by iteratively perturbing a locally optimal
    solution, next applying a local search algorithm to obtain a new
    locally optimal solution, and finally using an acceptance criterion
    for deciding from which of these two solutions to continue the
    search. The perturbation must be strong enough to allow the local
    search to escape from local optima and explore different areas
    of the search space, but also weak enough to avoid that the 
    algorithm degenerates into a simple random restart algorithm (that 
    typically performs poorly).

    \begin{figure}[t]
    \centering
    \begin{tabular}{@{} p{\textwidth} @{}}
    \hline
    \textbf{\emph{Iterative Local Search} (ILS)}\\
    \hline
    \vspace{0.0cm}

    \begin{minipage}[!t]{\textwidth}

    \hspace{1.5cm} $s_0$ = \verb GenerateInitialSolution (\emph{seed})

    \hspace{1.5cm} $s^*$ = \verb LocalSearch ($s_0$)

    \hspace{1.5cm} \textbf{repeat}

    \hspace{2cm} $s'$ = \verb Perturbation ($s^*$, \emph{history})

    \hspace{2cm} $s^*$$'$ = \verb LocalSearch ($s'$)

    \hspace{2cm} $s^*$ = \verb AcceptanceCriterion ($s^*$, $s^*$$'$, \emph{history})

    \hspace{1.5cm} \textbf{until} termination condition met

    \end{minipage}

    \vspace{0.2cm}\\
    \hline
    \end{tabular}
    \caption{Pseudo-code of Iterative Local Search (ILS).}
    \label{fig:ils}
    \end{figure}

    Figure \ref{fig:ils} depicts the four components that have to be
    specified to apply an ILS algorithm. The first one is the procedure
    \verb GenerateInitialSolution  that generates an initial solution
    $s_0$. The second one is the procedure \verb LocalSearch  that
    implements the local search algorithm, giving the mapping from a
    solution $s$ to a local optimal solution $s^*$. Any local search 
    algorithm can be used, however, the performance of the ILS algorithm 
    depends strongly on the one chosen. The \verb Perturbation  is
    responsible for perturbing the local optima $s^*$, returning a
    perturbed solution $s'$. Finally, the procedure
    \verb AcceptanceCriterion  decides which solution ($s^*$ or $s^*$$'$)
    will be perturbed in the next iteration. One important aspect of
    the perturbation and the acceptance criterion is to introduce a bias
    between intensification and diversification of the search. Intensification
    in the search can be reached by applying the perturbation always to
    the best solution found and using small perturbations. On the other hand, 
    diversification is achieved by accepting
    every new solution $s^*$$'$ and applying large perturbations.

\section{Two Search Models, Two Tracks, One Objective} \label{sec:ils+ecga}

  Different approaches have been proposed to combine global search with local
  search. A common practice is to combine GAs with local search heuristics. It 
  has been used so often that originated a new class of search methods called
  memetic algorithms \cite{Moscato:89}. In this work we propose something
  different, the combination of two global search methods based on distinct
  principles. By principles we mean variation operators, selection methods, and
  population management policies. The ECGA is a powerful search algorithm
  based on recombination to improve solutions, however at the cost of extra
  computation time (needed to search for a good MPM) in each generation. For
  hard problems this effort is well worth it, but for other problems, less
  complex search algorithms may do. This is where ILS comes in. As a
  light mutation-based algorithm, ILS can quickly and reliably find good 
  solutions for simpler or mutation-tailed problems.
  What we propose is to run ILS and ECGA simultaneously. This 
  ``pseudo-parallelism'' is done by giving an equal number of function 
  evaluations to each search method alternately. ILS and ECGA
  will have their own track in the exploration of the search
  space, without influencing each other. In the resulting
  optimization algorithm, that we call ILS+ECGA, the search will 
  be done by alternating between ILS and ECGA.

  \subsection{Parameter-less ECGA}

	The parameter-less GA technique is coupled together with the ECGA. 
	An important aspect of its implementation is the saving of 
	function evaluations. Since the crossover probability is always 
	equal to 0.5, there is no need of reevaluating the individuals 
	that are not sampled from the model. This way, half of the 
	total number of function
    	evaluations are saved. Note that the cost of running ILS
    	simultaneously is compensated by evaluating just half of the
    	individuals from the populations of the parameter-less ECGA.


  \subsection{ILS with Adaptive Perturbation}

    	In this section we describe the ILS implementation used for this 
	work, and present a simple but effective way to eliminate the 
	need of specifying its parameters. The four components chosen 
	for the ILS algorithm are:\\

    	\noindent \textbf{Local Search:}
    	next ascent hill climber (NAHC). NAHC consists in having one 
	individual and keep mutating each gene, one at a time, in a predefined 
	random sequence, until the resulting individual is fitter than the 
	original. In that case, the new individual replaces the 
	original and the procedure is repeated until no improvement 
	can be made further.\\ 

    	\noindent \textbf{Initial Solution:}
    	randomly generated. Since the NAHC is fast in getting local optima 
	solutions, there is no need to use a special greedy algorithm.\\

    	\noindent \textbf{Acceptance Criterion:}
	accept always the last local optima obtained ($s^*$$'$) 
	as the solution from where the search will continue. In a way, 
	this is done to compensate the intensive selection criterion from
    	NAHC, where just better solutions are accepted. On the other 
	side, with this kind of acceptance criterion we promote a
    	stochastic search in the space of local optima.\\

    	\noindent \textbf{Perturbation:}
    	probabilistic 
	and greater than the mutation rate of the NAHC (equal to $1/l$). 
	The perturbation strength is proportional to the problem size 
	(number of genes $l$). This way, the perturbation is always 
	strong enough, whatever the problem size. Each allele is 
	perturbed with probability $p_p = 0.05l/l = 0.05$. This means 
	that on average 5\% of the genes are perturbed. However, if 
	the problem length is too small (for example, $l \le 60$), then 
	the perturbation becomes of the same order of magnitude than 
	the mutation done by NAHC.
    	To avoid this, we fix the perturbation probability to
    	$3/l$ for problems where $l \le 60$. This way, we ensure that
    	on average the perturbation strength is at least 3 times
    	greater than the mutation strength of NAHC. This is done
   	to prevent perturbation from being easily canceled by the local 
	search algorithm. Nevertheless, the perturbation strength may not
    	be strong enough if the attraction area of a specific local optima is
    	too big, leading to a situation where frequently $s^* = s^*$$'$.
    	In that case, we need to increase the perturbation strength
    	until we get out from the attraction area of
    	the local optima. Therefore, the perturbation strength $\alpha$ 
	is updated as follows:

	\begin{equation}
        \alpha_{new} = \left\{ \begin{array}{ll}
              \alpha_{current} + 0.02l,    & \textrm{if $s^*$$'$ $=$ $s^*$} \\
              0.05l,   & \textrm{if $s^*$$'$ $\neq$ $s^*$}
              \end{array} \right.
        \end{equation}

 	\noindent
	This way, the updated perturbation probability is equal to 
	$\alpha_{new}/l$.


  \subsection{ILS+ECGA}

    	The parameter-less optimization framework proposed consists of 
	running the two different search models more or less simultaneously. 
	This is accomplished by switching back and forth between one method 
	and the other after a predefined number of function evaluations have 
	elapsed. Notice however that there are minimum execution units that 
	must be completed. For example, a generation of the parameter-less 
	ECGA cannot be left half done. Likewise, a NAHC search cannot be 
	interrupted in the middle. Therefore, care must be taken to ensure 
	that both methods receive approximately the same number of 
	evaluations and closest as possible from the defined value.
	For our experiments we used $fe_{elapsed}=500$.
	The ideal $fe_{elapsed}$ would be equal to one, 
	since the computational cost of changing between methods is 
	minimal. However, in practice, it will never happen because of 
	the minimal execution units of the ILS and ECGA. Since the main 
	objective of this work is to propose a parameter-less search 
	method, $fe_{elapsed}$ was fixed to a reasonable value. 

    	Initially, ILS with adaptive perturbation runs during 500 
	function evaluations, plus the ones needed to finish the current 
	NAHC search. Then, the parameter-less ECGA will run during another 
	500 evaluations, plus the ones needed to complete the 
	current generation. And the process repeats 
	\emph{ad eternum} until the user is satisfied with the solution 
	quality obtained or run out of time. This approach supplies
	robustness, small intervention from the user (just the fitness 
	function needs to be specified), and good results in a broad 
	class of problems. 

\section{Experiments} \label{sec:experiments}

  This section presents computer simulations on five test problems. These 
  problems were carefully chosen to represent different types of problem 
  difficulty. For each problem, the performance of ILS+ECGA algorithm is 
  compared with other four search algorithms: the simple GA with 
  ``standard'' parameters (SGA1), the 
  simple GA with tuned parameters (SGA2), the ILS with adaptive perturbation 
  alone (ILS), and the parameter-less ECGA alone (ECGA).
  For the GAs, we use binary encoding, tournament selection, 
  and uniform crossover (except for ECGA). SGA1 represents a typical GA 
  parameter configuration: population size $N=100$, crossover probability 
  $p_c=0.6$, mutation probability $p_m=0.001$, and selection pressure $s=2$. 
  SGA2 represents a tuned GA parameter configuration. For each problem, the 
  GA parameters were tuned to obtain the best performance. Note that they 
  aren't optimal parameters, but the best parameters found after a period 
  of wise trials\footnote{These trials were based on the work of Deb \& 
  Agrawal \cite{Deb:99}, since they used the same test functions. For 
  each trial, 5 runs were performed to get some 
  statistical significance.}. ILS and ECGA are tested alone to compare 
  with ILS+ECGA and understand the advantages of running the two 
  search models simultaneously.

  For each problem, 20 independent runs were performed in order to get
  results with statistical significance. For each run, 2,000,000 function 
  evaluations were allowed to be spent. For each algorithm, the mean and 
  standard deviation of the number of function evaluations spent to find 
  the target solution were calculated. For function optimization 
  testing, each run was considered well succeeded if it found a solution
  with a function value $f(x_1,\ldots,x_n)$ in a given neighborhood of the 
  optimal function value $f(x_{1opt},\ldots,x_{nopt})$. The number of runs 
  ($R_{ts}$) in which a target solution was found was also recorded. For 
  each problem, all algorithms started with the same 20 seed numbers in 
  order to avoid initialization (dis)advantages among algorithms.

  \subsection{Test Functions}

	The first problem is the onemax function, 
	that simply returns the number of ones in a string.
        A string length of 100 bits is used. The optimal solution is the
        string with all ones. After some tuning, SGA2 was set with $N=30$, 
	$p_{c}=0.9$, $p_{m}=0.005$, and $s=2$.

        The second test function is the unimodal Himmelblau's function, defined as
        $f(x_1,x_2) = ( x_{1}^2 + x_2 - 11 )^2 + (x_1 + x_{2}^2 - 7 )^2$. The
        search space considered is in the range $0 \leq x_1,x_2 \leq 6$, in which
        the function has a single minimum at (3,2) with a function value
        equal to zero. Each variable $x_i$ is
        encoded with 12 bits, totalizing a 24-bit chromosome. For a successful
        run, the function value must be smaller or equal to 0.001. After some
        tuning, SGA2 was set with the parameters $N=100$, $p_c=0.9$, $p_m=0.01$,
        and $s=2$.
                                                                                            
        The third function is the four-peaked Himmelblau's function,
        defined as $f(x_1,x_2) = ( x_{1}^2 + x_2 - 11 )^2 +
        (x_1 + x_{2}^2 - 7 )^2 + 0.1 (x_1-3)^2 (x_2-2)^2$. This function is
        similar to the previous one, but the range is extended to
        $-6 \leq x_1,x_2 \leq 6$. Since the original Himmelblau's function
        has four minima in this range (one in each quadrant), the added term
        causes the point (3,2) to be global minimum. 
	Each variable $x_i$ is
        encoded with 13 bits, giving a chromosome with 26 bits. Once more, a
        run is considered successful if the function value is within 0.001 of 
	the global optima. The SGA2 uses $N=200$, $p_c=0.5$, $p_m=1/l$, and $s=4$.
                                                                                            
        The fourth function tested is the 10-variable Rastrigin's
        function. This is a massively multimodal function, known to be
        difficult to any search algorithm. It is defined as $f(x_1,\ldots,x_{10})
        = 100 + \sum_{i=1}^{10} x_{i}^2 - 10 \cos (2 \pi x_i)$, being
        each variable defined in the range $-6 \leq x_i \leq 6$. This function has
        a global minimum at (0,0,...,0) with a function value equal to zero.
        There are a total of $13^{10}$ minima, of which $2^{10}$ are close to the
        global minimum. For best performance, SGA2 was set to $N=$ 10,000,
        $p_c=0.9$, $p_m=1/l$, and $s=8$.

	The fifth and last problem is a bounded deceptive function, that results
        from the concatenation of 10 copies of a 4-bit trap function.
        In a 4-bit trap function the fitness value depends on the number
        of ones ($u$) in a 4-bit string. If $u \leq 3$, the fitness is
        $3 - u$, if $u = 4$, the fitness is equal to $4$. The
        overall fitness is the sum of the 10 independent sub-function
        values. For such a problem, the SGA is only able to mix
        the building blocks with very large population sizes. To assure
        that we find the optimal solution in all 20 runs, the SGA2
        was set with $N$= 60,000, $p_c=0.5$, $p_m=0$, and $s=4$.

	\subsection{Results}

        \begin{table}[!t]
        \caption{Mean and standard deviation of the number of function evaluations
                 spent to find the target solution for the tested problems. The 
		 number of runs ($R_{ts}$) in which a target solution was found was also 
		 recorded.}
        \label{tab:artificial}
        \begin{center}
        \begin{tabular}{c|c c c c c c }
        \hline
       & \hspace{1.4cm} & \textbf{SGA1} & \textbf{SGA2} & \textbf{ECGA}
            & \textbf{ILS} & \textbf{ILS+ECGA} \\
        \hline
                & mean       & 2,990 & 1,256 & 13,735 & 451 &    451  \\
\textbf{Onemax} & std. dev.  & $\pm$189 & $\pm$258 & $\pm$5,371 & $\pm$65 & $\pm$65 \\
                & $R_{ts}$ &   20  &   20  &   20   &  20 &  20+0 \\
        \hline
\textbf{Unimodal}   & mean       & 2,019 & 1,750 & 3,731 & 1,400 & 3,174  \\
\textbf{Himmelblau} & std. dev.  & $\pm$790 & $\pm$497 & $\pm$2,290
            & $\pm$1,385 & $\pm$2,766  \\
                    & $R_{ts}$ &   16  &   20  &   20  &  20   & 14+6\\
        \hline
\textbf{Four-peaked} & mean       & 2,414 & 2,850 & 5,205 & 2,593 & 4,990  \\
 \textbf{Himmelblau} & std. dev.  & $\pm$750 & $\pm$668 & $\pm$2,725
             & $\pm$3,002 & $\pm$3,432  \\
                     & $R_{ts}$ &   14  &   20  &   20  &  20   & 12+8\\
        \hline
\textbf{10-variable} & mean       & 1,555,300 & 570,000 & 149,635 & $>$2,000,000
             & 275,170\\
  \textbf{Rastrigin} & std. dev.  & $\pm$306,600 & $\pm$87,240 & $\pm$85,608
             & --- & $\pm$87,472\\
                     & $R_{ts}$ &   3  &   20  &   20  &  0  & 0+20\\
        \hline
   \textbf{Bounded} & mean       & ---   & 741,000 & 15,388
              & $>$2,000,000 & 31,870 \\
 \textbf{Deceptive} & std. dev.  & ---   &   $\pm$95,416 &  $\pm$3,417
              &    ---     & $\pm$15,306 \\
                    & $R_{ts}$ &   0   &   20      &   20
              &  0  & 0+20\\
        \hline
        \end{tabular}
        \end{center}
        \end{table}

	The results obtained can be seen in Table \ref{tab:artificial}. 
	The growing difficulty of the five tested problems
        can be verified by the number of runs ($R_{ts}$) in which algorithms found a
        target solution, and by the number of function evaluations needed 
	to do so. For the onemax problem, all the algorithms found the target 
	solution in the 20 runs. Both ILS and ILS+ECGA got the same (and the
        best) results. This can be explained because the ILS is the
        first search method to run (500 function evaluations) in the ILS+ECGA
        framework. Taking into account that both algorithms used the same seed
        numbers, it was expected that they did similar since the NAHC always
        returned the optimal solution in the first time that it was solicited.
        This eventually happens because the problem is linear in Hamming
        space. In fact, that's the reason why mutation-based algorithms
        outperformed the rest of the algorithms for this problem.                                     For the remaining problems, the SGA1 (with ``standard'' parameters) 
	couldn't find a satisfiable solution in all runs. Although SGA1 
	performed well for the Himmelblau's functions, it wasn't robust enough 
	to achieve a target solution in all runs. For the 
	10-variable Rastrigin's function, SGA1 found only 3 good solutions, 
	and for the deceptive function, the ``standard'' parameter configuration 
	failed completely, converging always to sub-optimal solutions. These 
	results confirmed that setting these parameters to all kind of problems 
	is a mistake. 

	For the Himmelblau's functions, SGA2 and ILS obtained the best results, 
	taking half of the evaluations spent by ECGA. The ILS+ECGA algorithm, 
	mostly due to the ILS performance, obtained a good performance. 
	Note that ILS was the algorithm 
	responsible for getting a good solution in 14 and 12 (in a total of 20) 
	runs, for unimodal and four-peaked Himmelblau's functions, respectively.

	For the 10-variable Rastrigin's function, a different scenario occurred. 
	ILS failed all the attempts to find a satisfiable solution. 
	This is not a surprising result, since 
	search algorithms based on local search don't do well in massively 
	multimodal functions. Remember that some of the components (NAHC and 
	adaptive perturbation scheme) of our ILS implementation were chosen 
	in order to solve linear, non-correlated, or mutation-tailed problems 
	in a quick and reliable way. For other kind of problems, 
	parameter-less ECGA performance is quite good, making ILS+ECGA a 
	robust and easy-to-use search algorithm. The ECGA was the best 
	algorithm for this problem, and because of it, ILS+ECGA got the second 
	best result, taking half of the evaluations of the SGA2. 

	For the bounded deceptive problem, SGA1 (converged to sub-optimal 
	solutions) and ILS (spent all of the 2,000,000 evaluations available) 
	didn't find the optimal solution.
	For this problem, the real power of ECGA could be verified. 
	SGA2 took almost 50 more times function evaluations than
        ECGA to find the best solution in all runs. Taking advantage
        of the ECGA performance, ILS+ECGA was the second best algorithm, finding 
	the target solution in 2 times more evaluations than the ECGA alone. 

\section{Extensions} \label{sec:extensions}

	There are a number of extensions that can be done based on this work:
	
	\begin{itemize}
	\item investigate other workload strategies.
	\item investigate interactions between the two search methods.
	\item investigate how other algorithms perform in the framework.
	\end{itemize}

	For many problems, the internal mechanisms needed by the ECGA to 
	build the MPM may contribute to a significant fraction of the total 
	execution time. Therefore, it makes sense (and it's more fair) to divide 
	the workload between the two methods based on total execution time 
	rather than on fitness function evaluations. Another aspect is to 
	investigate interactions between the two methods. How much beneficial 
	is it to insert one (or more) ILS local optimal solution(s) in one 
	(or more) population(s) of the parameter-less ECGA? What about the 
	reverse situation? Finally, other algorithm instances such as BOA 
	could be used instead of the ECGA, as well as other concrete ILS 
	implementation.

	We are currently exploring some of these extensions.

\section{Summary and Conclusions} \label{sec:conclusions}

	This paper presented a concrete implementation of the proposed 
	parameter-less optimization framework that 
	eliminates the need of specifying the configuration parameters, 
	and combines population-based search with iterated local search 
	in a novel way. The user just needs to specify the fitness function 
	in order to achieve good solutions for the optimization problem.
	
	Although the combination might not perform as well as the best algorithm 
	for a specific problem, it is more robust than either method alone, 
	working reasonably well on problems with different characteristics.
 


\bibliography{references}

\begin{thebibliography}{10}

\bibitem{Spears:93}
Spears, W.M.:
\newblock Crossover or mutation?
\newblock In Whitley, L.D., ed.: Foundations of Genetic Algorithms 2.
\newblock Morgan Kaufmann, San Mateo, CA (1993)  221--237

\bibitem{DeJong:75}
{De Jong}, K.A.:
\newblock An analysis of the behavior of a class of genetic adaptive systems.
\newblock PhD thesis, University of Michigan, Ann Arbor (1975)

\bibitem{Grefenstette:86}
Grefenstette, J.J.:
\newblock Optimization of control parameters for genetic algorithms.
\newblock In Sage, A.P., ed.: IEEE Transactions on Systems, Man, and
  Cybernetics. Volume SMC--16(1),.
\newblock IEEE, New York (1986)  122--128

\bibitem{Schaffer:89}
Schaffer, J.D., Caruana, R.A., Eshelman, L.J., Das, R.:
\newblock A study of control parameters affecting online performance of genetic
  algorithms for function optimization.
\newblock In Schaffer, J.D., ed.: Proceedings of the Third International
  Conference on Genetic Algorithms, San Mateo, CA, Morgan Kaufman (1989)
  51--60

\bibitem{Goldberg:91}
Goldberg, D.E., Deb, K.:
\newblock A comparative analysis of selection schemes used in genetic
  algorithms.
\newblock Proceedings of the First Workshop on Foundations of Genetic
  Algorithms \textbf{1} (1991)  69--93 (Also {TCGA R}eport 90007).

\bibitem{Goldberg:92}
Goldberg, D.E., Deb, K., Clark, J.H.:
\newblock Genetic algorithms, noise, and the sizing of populations.
\newblock Complex {S}ystems \textbf{6} (1992)  333--362

\bibitem{Harik:97a}
Harik, G., Cantú-Paz, E., Goldberg, D.E., Miller, B.L.:
\newblock The gambler's ruin problem, genetic algorithms, and the sizing of
  populations.
\newblock In: Proceedings of the {I}nternational {C}onference on {E}volutionary
  {C}omputation 1997 (ICEC '97), Piscataway, NJ, {IEEE} {P}ress (1997)  7--12

\bibitem{Muhlenbein:92}
M{\"{u}}hlenbein, H.:
\newblock How genetic algorithms really work: {I}.{M}utation and
  {H}illclimbing.
\newblock In M{\"{a}}nner, R., Manderick, B., eds.: {P}arallel {P}roblem
  {S}olving from {N}ature 2, Amsterdam, The Netherlands, Elsevier Science
  (1992)  15--25

\bibitem{Back:93a}
B{\"{a}}ck, T.:
\newblock Optimal mutation rates in genetic search.
\newblock In: Proceedings of the Fifth International Conference on Genetic
  Algorithms. (1993)  2--8

\bibitem{Goldberg:93}
Goldberg, D.E., Deb, K., Thierens, D.:
\newblock Toward a better understanding of mixing in genetic algorithms.
\newblock Journal of the Society of Instrument and Control Engineers
  \textbf{32} (1993)  10--16

\bibitem{Thierens:93}
Thierens, D., Goldberg, D.E.:
\newblock Mixing in genetic algorithms.
\newblock In: Proceedings of the Fifth International Conference on Genetic
  Algorithms. (1993)  38--45

\bibitem{Eiben:99}
Eiben, A.E., Hintering, R., Michalewicz, Z.:
\newblock Parameter {C}ontrol in {E}volutionary {A}lgorithms.
\newblock {IEEE} {T}ransactions on {E}volutionary {C}omputation \textbf{3}
  (1999)  124--141

\bibitem{Davis:89}
Davis, L.:
\newblock Adapting operator probabilities in genetic algorithms.
\newblock In Schaffer, J.D., ed.: Proceedings of the Third International
  Conference on Genetic Algorithms, San Mateo, CA, Morgan Kaufman (1989)
  61--69

\bibitem{Julstrom:95}
Julstrom, B.A.:
\newblock What have you done for me lately? {A}dapting operator probabilities
  in a steady-state genetic algorithm.
\newblock In Eshelman, L., ed.: Proceedings of the Sixth International
  Conference on Genetic Algorithms, San Francisco, CA, Morgan Kaufmann (1995)
  81--87

\bibitem{Smith:95}
Smith, R.E., Smuda, E.:
\newblock Adaptively resizing populations: Algorithm, analysis, and first
  results.
\newblock Complex Systems \textbf{9} (1995)  47--72

\bibitem{Back:95}
B{\"{a}}ck, T., Schwefel, H.P.:
\newblock Evolution strategies {I}: {V}ariants and their computational
  implementation.
\newblock In Winter, et~al., eds.: Genetic Algorithms in Engineering and
  Computer Science.
\newblock John {W}iley and {S}ons, Chichester (1995)  111--126

\bibitem{Harik:99}
Harik, G.R., Lobo, F.G.:
\newblock A parameter-less genetic algorithm.
\newblock In Banzhaf, et~al., eds.: Proceedings of the {G}enetic and
  {E}volutionary {C}omputation {C}onference {GECCO}-99, San Francisco, CA,
  Morgan Kaufmann (1999)  258--265

\bibitem{Holland:75}
Holland, J.H.:
\newblock Adaptation in Natural and Artificial Systems.
\newblock University of Michigan Press, Ann Arbor, MI (1975)

\bibitem{Harik:99a}
Harik, G.R.:
\newblock Linkage learning via probabilistic modeling in the {ECGA}.
\newblock Illi{GAL} Report No. 99010, Illinois Genetic Algorithms Laboratory,
  University of Illinois at Urbana-Champaign, Urbana, IL (1999)

\bibitem{Lourenco:02}
Lourenço, H.R., Martin, O., St{\"{u}}tzle, T.:
\newblock Iterated local search.
\newblock In Glover, F., Kochenberger, G., eds.: Handbook of Metaheuristics,
  Norwell, MA, Kluwer Academic Publishers (2002)  321--353

\bibitem{Moscato:89}
Moscato, P.:
\newblock On evolution, search, optimization, genetic algorithms and martial
  arts: Towards memetic algorithms.
\newblock Technical Report C3P 826, Caltech {C}oncurrent {C}omputation
  {P}rogram, {C}alifornia {I}nstitute of {T}echnology, Pasadena, CA (1989)

\bibitem{Deb:99}
Deb, K., Agrawal, S.:
\newblock Understanding interactions among genetic algorithm parameters.
\newblock In Banzhaf, W., Reeves, C., eds.: Foundations of Genetic Algorithms 5
  (FOGA'98), Amsterdam, Morgan Kaufmann, San Mateo CA, 1999 (1998)  265--286

\end{thebibliography}

\end{document}